\documentclass[letterpaper]{article} 
\usepackage{aaai2026}  
\usepackage{times}  
\usepackage{helvet}  
\usepackage{courier}  
\usepackage[hyphens]{url}  
\usepackage{graphicx} 
\usepackage{natbib}  
\usepackage{caption} 
\usepackage{algorithm}
\usepackage{algorithmic}
\usepackage{amsmath}
\usepackage{amssymb}
\usepackage{xcolor}
\usepackage{epstopdf}
\usepackage{booktabs}

\usepackage{newfloat}
\usepackage{listings}
\DeclareCaptionStyle{ruled}{labelfont=normalfont,labelsep=colon,strut=off} 
\floatstyle{ruled}
\newfloat{listing}{tb}{lst}{}
\floatname{listing}{Listing}
\nocopyright 

\title{StyDeco: Unsupervised Style Transfer with \\ Distilling Priors and Semantic Decoupling}
\author {
    Yuanlin Yang\equalcontrib,
    Quanjian Song\equalcontrib,
    Zhexian Gao,
    Ge Wang,
    Shanshan Li,
    Xiaoyan Zhang
}

\usepackage{bibentry}

\begin{document}

\maketitle

\begin{abstract}
Diffusion models have emerged as the dominant paradigm for style transfer,  but their text-driven mechanism is hindered by a core limitation: it treats textual descriptions as uniform, monolithic guidance.
This limitation overlooks the semantic gap between the non-spatial nature of textual descriptions and the spatially-aware attributes of visual style, often leading to the loss of semantic structure and fine-grained details during stylization.
In this paper, we propose StyDeco, an unsupervised framework that resolves this limitation by learning text representations specifically tailored for the style transfer task.
Our framework first employs Prior-Guided Data Distillation (PGD), a strategy designed to distill stylistic knowledge without human supervision. It leverages a powerful frozen generative model to automatically synthesize pseudo-paired data.
Subsequently, we introduce Contrastive Semantic Decoupling (CSD), a task-specific objective that adapts a text encoder using domain-specific weights.
CSD performs a two-class clustering in the semantic space, encouraging source and target representations to form distinct clusters.
Extensive experiments on three classic benchmarks demonstrate that our framework outperforms several existing approaches in both stylistic fidelity and structural preservation, highlighting its effectiveness in style transfer with semantic preservation.
In addition, our framework supports a unique de-stylization process, further demonstrating its extensibility.
Our code is vailable at \url{https://github.com/QuanjianSong/StyDeco}.
\end{abstract}


\section{Introduction}
Style transfer~\cite{Parmar2023ZeroshotIT,Huang2023SmartEditEC,song2024univst} is a fundamental task in computer vision that renders an input image with a given style while preserving its essential spatial structures. This technique plays a vital role in various inspiring applications, such as artistic creation and film production.
Traditionally, performing this task relies heavily on manual effort, requiring professional artists to take the source image as a starting point and create artwork based on it, making the process labor-intensive and time-consuming.
With the rise of deep learning, style patterns can now be learned automatically from large-scale datasets.
Early approaches commonly adopted generative models like GANs~\cite{Goodfellow2021GenerativeAN}, particularly CycleGAN~\cite{Zhu2017UnpairedIT}, which performs unsupervised stylization and naturally suits the scarcity of real style images.
However, due to the unstable training and limited generative capacity, GAN-based methods struggle to generalize to broader real-world applications.

\begin{figure}[t]
    \centering
    \includegraphics[width=\linewidth]{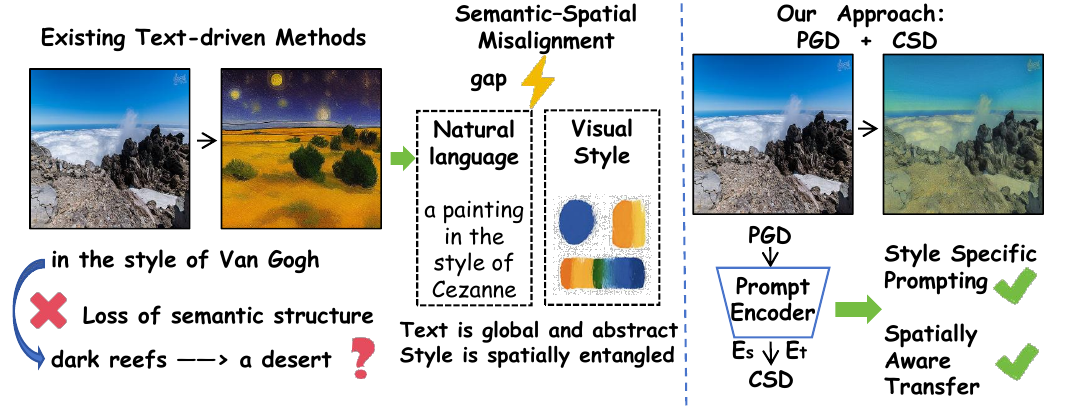} 
    \caption{Graphical illustration of the main distinctions between existing methods (left) and StyDeco (right).}
    \label{fig:motivation}
\end{figure}

Recent advancements in diffusion models have driven a paradigm shift in generative AI, marked by their superior generative capacity and training stability compared to prior GAN-based approaches.
Powered by advances in multimodal learning, text-driven architectures such as the Stable Diffusion series~\cite{Rombach2021HighResolutionIS} and FLUX have become dominant frameworks, laying the foundation for downstream tasks like style transfer.
Existing style transfer methods are generally categorized into three types: instruction-based, training-based, and training-free approaches.
\textit{(i) Instruction-based style transfer.}
These methods represent a general form of style transfer, where image-to-image translation is guided by natural language instructions. These methods typically rely on conditional models such as instruction-tuned diffusion models~\cite{Brooks2022InstructPix2PixLT} and multimodal large language models~\cite{Fu2023GuidingII} to capture the instruction-following behavior. Recent work further improves this process by incorporating visual reasoning~\cite{Huang2023SmartEditEC}.
\textit{(ii) Training-based style transfer.}
These methods mainly learn the image-to-image mapping through large-scale conditional fine-tuning. Representative studies construct this mapping using one-step generators~\cite{Parmar2024OneStepIT}, textual representation learning~\cite{Zhang2022InversionbasedST}, or the Schrödinger Bridge framework~\cite{Su2022DualDI}.
\textit{(iii) Training-free style transfer.}
These methods aim to fully leverage pre-trained generative backbone models. They typically modify the sampling process to enforce structure preservation from different perspectives, such as spatial features~\cite{Tumanyan2022PlugandPlayDF}, cross-attention computation~\cite{Deng2023ZZS}, and the text embedding space~\cite{Parmar2023ZeroshotIT}.

However, this reveals a flawed assumption underlying current text-driven methods: they either attempt to map explicit linguistic descriptions to visual style or rely on black-box, deeply customized style generation.
Both approaches fundamentally overlook the fact that natural language semantics are inherently non-spatial, whereas visual style is typically entangled with spatially aware attributes.
As shown in Figure~\ref{fig:motivation}, applying direct text prompts to spatially sensitive tasks often results in semantic structure loss in stylized outputs, as such prompts impose abstract and uniform semantics on processes that require task-specific and spatially coherent guidance.
This mismatch between conventional prompting and task’s spatial demands motivates our work to develop stylistically-aware text prompts that better preserve spatial details for style transfer.

In this paper, we propose \textbf{StyDeco}, an unsupervised style transfer framework that integrates distilled priors and semantic decoupling to address the aforementioned challenges.
Our framework builds upon the power of generative priors and introduces a two-stage learning scheme comprising \textit{Prior-Guided Data Distillation (PGD)} and \textit{Contrastive Semantic Decoupling (CSD)}, aiming to bridge the gap between textual descriptions and spatially coherent visual styles.
Leveraging a powerful off-the-shelf generator, PGD distills stylized data using specific template prompts, which is then used for unsupervised style transfer training.
With the help of distilled knowledge, CSD explicitly learns the semantic differences between the source and target domains. It introduces two distinct sets of projection parameters to map domain-specific prompts into separate clusters, within the latent semantic space of the text encoder. This design enables adaptive, task-specific encoding for unsupervised style transfer.
Extensive experiments on the applications of three representative style patterns, including Van Gogh, Cezanne, and Ukiyoe, show that our method outperforms several existing approaches.
In addition, our framework supports a unique de-stylization function, further highlighting its distinct advantages over existing methods.



\begin{figure*}[htbp]
    \centering
    \includegraphics[width=\textwidth]{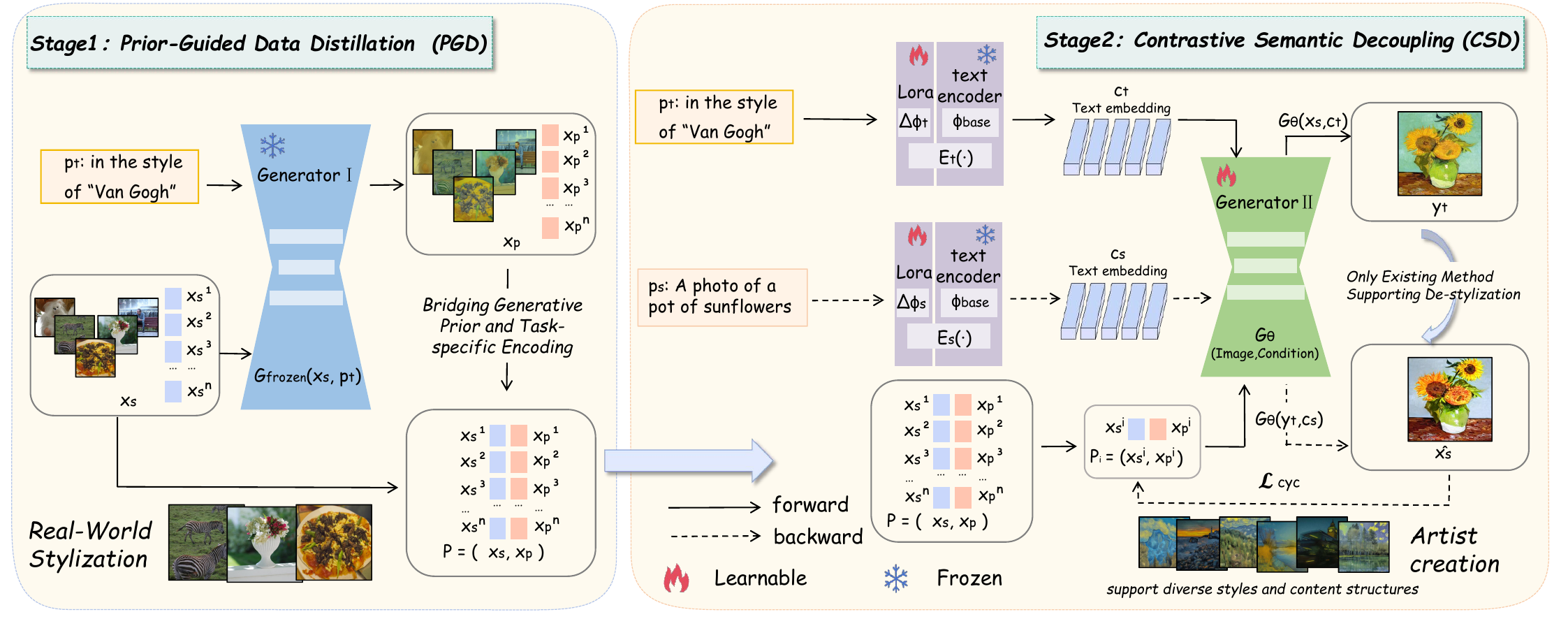} 
    \caption{
    An overview of the framework of our proposed StyDeco, which consists of two primary stages.
    In Prior-Guided Data Distillation (PGD), we leverage a powerful frozen generator $G_{frozen}$ to synthesize stylized images $x_p$ from a collection of natural images and a target style prompt $p_t$.
    In Contrastive Semantic Decoupling (CSD), we introduce two distinct LoRA-adapted text encoders $E_s$ and $E_t$, to learn decoupled embeddings for the content and style domains respectively.
    These specialized encoders guide our main generator $G_\theta$ within a cycle-consistent training loop, where a cycle loss $\mathcal{L}_{cyc}$ ensures the fidelity of reconstruction in the absence of paired supervision.
    }
    \label{fig:framework}
\end{figure*}

\section{Related Work}

Style transfer has long been a popular research topic in computer vision.
Early studies~\cite{Chen2017CoherentOV,Isola2016ImagetoImageTW,Sheng2018AvatarNetMZ,Kolkin2019StyleTB,Baek2020RethinkingTT} primarily address the style transfer task using explicit supervision signals, emphasizing sample quality in specific scenarios with paired data or conditional labels.
However, these conventional stylized tasks are normally data-hungry, which fundamentally conflicts with the fact that real artistic images are tough to collect, especially the ones that are paired with natural images.
As a result, supervised style transfer methods face inherent limitations in real-world applications.
To overcome these challenges, unsupervised methods~\cite{Choi2017StarGANUG,Kim2017LearningTD,Huang2018MultimodalUI,Lee2018DiverseIT} have attracted increasing attention.
These methods loose the requirement of paired training in conventional paradigms of style transfer, where they only require two sets of images to automatically learn the internal mapping.
Representative studies like CycleGAN~\cite{Zhu2017UnpairedIT} employs two generators to achieve cycle-consistent mappings between two separated domains, yet GAN-based training is often unstable and thus prevents further improvements.
Recently, diffusion model-based approaches~\cite{Song2020ImprovedTF,Ho2020DenoisingDP,Song2020ScoreBasedGM,Rombach2021HighResolutionIS,Saharia2022PhotorealisticTD,Zhang2023AddingCC,song2024univst} have gradually become predominant in image-to-image translation due to their improved training stability and generation quality.
However, most existing diffusion models primarily focus on preserving image content and structure, while paying relatively little attention to leveraging the potentially beneficial connections between text and style.
This limitation is especially evident in text-driven style transfer~\cite{Patashnik2021StyleCLIPTM,Nichol2021GLIDETP,Ramesh2022HierarchicalTI,Zhang2023AddingCC}, where such potential remains largely underexplored.
Although alternative solutions have been proposed using either tuning-based~\cite{Brooks2022InstructPix2PixLT, Fu2023GuidingII, Huang2023SmartEditEC} or tuning-free methods~\cite{Tumanyan2022PlugandPlayDF, Deng2023ZZS, Parmar2023ZeroshotIT}, they still struggle in scenarios that require a proper balance between semantic meaning and spatial structure in the stylized results.
This challenge motivates our work to address the issue more effectively.

\section{Method}

\subsection{Overall workflow}
Before we start introducing the details of components in our method, we first illustrate the overview of the unsupervised style transfer task.
Specifically, we build our overall framework based on Image-to-Image Turbo~\cite{Parmar2024OneStepIT}, a CycleGAN-style unsupervised paradigm based on a one-step diffusion model~\cite{Sauer2023AdversarialDD}.
Given a source domain of natural images $\mathcal{D}_s$ and a target domain $\mathcal{D}_t$ defined by a specific style, our goal is to learn a mapping function $G_{\theta}$ to convert $x\in \mathcal{D}_s$ to $y \in \mathcal{D}_t$, with such process conditioned on the textual description of the target style.
Figure~\ref{fig:framework} shows the overall pipeline of our method, consisting of two main stages: \textit{Prior-Guided Data Distillation (PGD)}, which synthesizes a domain translation dataset with pseudo supervision; and \textit{Contrastive Semantic Decoupling (CSD)}, which learns distinct semantic representations for both domains.

\subsection{Prior-Guided Data Distillation (PGD)}
\label{sec:pgd}
To enable unsupervised style transfer, our primary goal is to construct the foundational data necessary for standard cycle-consistent training.
While conventional approaches typically rely on large-scale unpaired datasets to support this process, such a paradigm proves challenging for style transfer, where style images are often scarce and difficult to obtain.
Fortunately, text-driven generative models demonstrate remarkable capability in addressing this limitation. With only simple text descriptions (e.g., ``a painting in the style of Van Gogh''), these models can synthesize high-quality, photorealistic images.
Inspired by this, we propose \textit{Prior-Guided Data Gistillation (PGD)}, which leverages such strong generative priors to facilitate data construction.
Specifically, we set the inputs of this stage as a collection of source images $\{\mathbf{x}_s\} \subset \mathcal{D}_s$ and a fixed text prompt template $p_t$ that describes the target style (e.g., ``\textit{in the style of Ukiyoe}'').
We freeze the model parameters of the generative prior, termed as $G_{frozen}$ (e.g., InstructPix2Pix~\cite{Brooks2022InstructPix2PixLT}) to act as a knowledge distiller.
Given each source image $\mathbf{x}_s$ in our collection $\mathcal{D}_s$, we perform a single forward pass through this frozen model to synthesize its stylized counterpart, which we denote as the pseudo-target image $\mathbf{x}_p$.
This process can be formally by:
\begin{equation}
    \mathbf{x}_p = G_{frozen}(\mathbf{x}_s, p_t).
    \label{eq:pgd}
\end{equation}
By iterating through the entire source collection, we effectively distill the implicit stylistic knowledge stored within the foundation model into an explicit dataset $\mathcal{P} = \{(\mathbf{x}_s, \mathbf{x}_p)\}$.
This curated dataset $\mathcal{P}$ provides the basic mapping relationship with pseudo-supervision signals, which are later utilized in the training stage of our model.

\subsection{Contrastive Semantic Decoupling (CSD)}
\label{sec:csd}
Once we curate the training data $\mathcal{P}$ for our method, the next step is to train our generator with the distilled knowledge.
Given that diffusion models function primarily as text-driven generators, it becomes crucial to carefully design the use of text prompts in the context of unsupervised style transfer. Existing approaches typically rely on explicit textual descriptions to represent the target style, yet they lack task-specific adaptation. In particular, they often fail to account for the entanglement between semantic features and spatially-aware details that are essential for high-quality stylization.
To address this, we introduce \textit{Contrastive Semantic Decoupling (CSD)}, a learning paradigm to tailor text encoding process for unsupervised style transfer.

\paragraph{Forward Processes.}
Given previous pseudo-paired data $(\mathbf{x}_s, \mathbf{x}_p) \in \mathcal{P}$, a source domain prompt $p_s$, and the target domain prompt $p_t$, we introduce a learning mechanism designed to resolve the conflict between non-spatial text and spatially-aware style.
We observe that widely adopted text encoders, such as the CLIP text encoder~\cite{Radford2021LearningTV} and T5~\cite{Raffel2019ExploringTL}, are insufficient for modeling the distinct semantics of different domains, due to the fact that their original objectives are not directly designed for image stylization.
An intuitive way to address this limitation is to directly integrate the optimization of text representations into the training process of the style generator, prompting the emergence of our CSD method.
In CSD, we utilize a frozen base text encoder $E_{base}$, with its original parameters $\phi_{base}$.
Instead of appending new modules, we introduce two lightweight, domain-specific sets of Low-Rank Adaptation (LoRA)~\cite{Hu2021LoRALA} weights, denoted as $\Delta\phi_s$ and $\Delta\phi_t$.
These LoRA weights adaptively modify the behavior of the base encoder for each domain, creating two specialized text encoders $E_s$ and $E_t$, formally expressed as:
\begin{align}
    E_s(\cdot) &= E_{base}(\cdot; \phi_{base} + \Delta\phi_s), \\
    E_t(\cdot) &= E_{base}(\cdot; \phi_{base} + \Delta\phi_t).
    \label{eq:lora_encoders}
\end{align}
The trainable parameters of our text encoding process are therefore only the low-rank weights $\{\Delta\phi_s, \Delta\phi_t\}$.

Following a cycle-consistent training paradigm of Image-to-Image Turbo~\cite{Parmar2024OneStepIT}, our framework performs a complete translation loop.
For the forward translation from the domain of natural image to the stylized ones, we first encode the target prompt $p_t$ using the target-specific encoder $E_t$ to obtain the style embedding $\mathbf{c}_t = E_t(p_t)$.
Then, our generator $G_{\theta}$ takes the source image $\mathbf{x}_s$ and this style embedding $\mathbf{c}_t$ to produce the stylized output $\mathbf{y}_t$:
\begin{equation}
    \mathbf{y}_t = G_{\theta}(\mathbf{x}_s, \mathbf{c}_t).
    \label{eq:forward_pass}
\end{equation}
For the backward translation in reverse, we use the generated image $\mathbf{y}_t$ as input, by encoding the source prompt $p_s$ with its corresponding specialized encoder $E_s$ to get the source embedding $\mathbf{c}_s = E_s(p_s)$.
The style-to-natural image generator $G_{\theta}$ then attempts to reconstruct the original source image:
\begin{equation}
    \hat{\mathbf{x}}_s = G_{\theta}(\mathbf{y}_t, \mathbf{c}_s).
    \label{eq:backward_pass}
\end{equation}
Similarly, we perform the same cycle loop given a style image as input, where we first project it to the domain of natural images, and ultimately convert it back to end up the loop.
For both loops given different inputs, we utilize the training objectives described in the following texts.

\paragraph{Training Objectives.}
The training of our model, including the generator weights $\theta$ and LoRA weights $\{\Delta\phi_s, \Delta\phi_t\}$, is guided by a composite loss function.
To learn the domain-to-domain mapping process as our base model does, we employ a cycle-consistency loss between the reconstructed image $\hat{\mathbf{x}}_s$ and the original source image $\mathbf{x}_s$, as formulated by:
\begin{equation}
    \mathcal{L}_{cyc} = \mathbb{E}_{\mathbf{x}_s \sim \mathcal{D}_s} \left[ ||\hat{\mathbf{x}}_s - \mathbf{x}_s||_1 \right].
    \label{eq:loss_cyc}
\end{equation}
Here, we typically use a combination of L1 and LPIPS losses~\cite{Zhang2018TheUE} to improve visual quality.


\paragraph{Inference Process.}
Once the model is trained, inference is a single, efficient forward pass.
To stylize a new source image $\mathbf{x}_{new}$, we first compute the target style embedding $\mathbf{c}_t = E_t(p_t)$ using the target-adapted encoder.
The final stylized image $\mathbf{y}_{final}$ is then generated by the trained generator $G_{\theta}$ as $\mathbf{y}_{final} = G_{\theta}(\mathbf{x}_{new}, \mathbf{c}_t)$.
This process is deterministic and fast, making it ideal for practical applications.


\section{Experiment}

\subsection{Experimental settings}

\paragraph{Implementation Details.}
We train four models on three different datasets based on SD-Turbo\footnote{\url{https://huggingface.co/stabilityai/sd-turbo}}, a distilled version of few-steps diffusion model inherited from the Stable Diffusion series~\cite{Rombach2021HighResolutionIS}.
The overall pipeline is trained with a total of $40,000$ steps and a batch size of $2$, where we particularly use the gradient accumulation techinques to save GPU memory consumption.
For optimization, we use the AdamW optimizer~\cite{Loshchilov2017DecoupledWD} with $\beta = (0.9, 0.999)$, a learning rate of $1\mathrm{e}{-5}$, and a weight decay of $1\mathrm{e}{-2}$.
We follow Image-to-Image Turbo \cite{Parmar2024OneStepIT} to inject Low-Rank (LoRA) adaptation matrices to the VAE and U-net of the diffusion model.
During training, we apply gradient clipping with a threshold of $1.0$ to ensure stability, and adopt mixed-precision to reduce memory usage.
All input images are resize to a resolution of $256 \times 256$.
All experiments are conducted on an NVIDIA A800-80GB GPUs with a fixed random seed of $42$.

\begin{table*}[t!]
\centering
\setlength{\tabcolsep}{0.3em}
\scalebox{1}{\begin{tabular}{@{}lcccccccccccc@{}}
\toprule
& \multicolumn{4}{c}{Van Gogh dataset} & \multicolumn{4}{c}{Cezanne dataset} & \multicolumn{4}{c}{Ukiyoe dataset} \\
\cmidrule(lr){2-5} \cmidrule(lr){6-9} \cmidrule(lr){10-13}
Method & FID$\downarrow$ & SSIM$\uparrow$ & LPIPS$\downarrow$ & CLIP-ac$\uparrow$ & FID$\downarrow$ & SSIM$\uparrow$ & LPIPS$\downarrow$ & CLIP-ac$\uparrow$ & FID$\downarrow$ & SSIM$\uparrow$ & LPIPS$\downarrow$ & CLIP-ac$\uparrow$ \\
\midrule
StyTr           & \textbf{162.8}    & 0.612             & 0.413             & 5.365             & \textbf{207.8}    & 0.614             & 0.401             & 5.294             & 218.6             & 0.584             & 0.409             & 5.160 \\
InstructPix2Pix & 193.8             & 0.398             & 0.435             & \textbf{5.934}    & 237.9             & 0.411             & 0.330             & \underline{5.562} & \underline{241.8} & 0.400             & 0.417             & 5.277 \\
Z*              & 195.2             & \underline{0.738} & \underline{0.277} & 5.330             & \underline{220.4} & \underline{0.738} & \underline{0.300} & 5.285             & 248.1             & \textbf{0.718}    & \textbf{0.310}    & \underline{5.334} \\
Puff-Net        & 299.9             & 0.331             & 0.467             & 4.304             & 318.6             & 0.332             & 0.467             & 4.304             & 287.8             & 0.332             & 0.467             & 4.304 \\
Ours            & \underline{188.2} & \textbf{0.813}    & \textbf{0.325}    & \underline{5.705} & 234.2             & \textbf{0.806}    & \textbf{0.199}    & \textbf{5.673}    & \textbf{244.5}    & \underline{0.608} & \underline{0.388} & \textbf{5.042} \\
\bottomrule
\end{tabular}}
\caption{Quantitative comparison with various methods on Van Gogh, Cezanne and Ukiyoe datasets. Best are in \textbf{bold}, second-best \underline{underlined}. Our method achieves superior overall performance.}
\label{tab:combined_results_all}
\end{table*}

\paragraph{Dataset Details.}
We conduct training and evaluation considering three main style patterns, including Van Gogh, Cezanne, and Ukiyoe.
To construct the training set, we randomly sample $5,000$ natural images from the COCO 2014 training set \cite{coco}; and follow our stylization procedures in PGD to obtain the style images.
For evaluation, we adopt three widely used datasets in style transfer: VanGogh2Photo\footnote{\url{https://huggingface.co/datasets/huggan/vangogh2photo/tree/main}}, Cezanne2Photo\footnote{https://huggingface.co/datasets/huggan/cezanne2photo}, and Ukiyoe2Photo\footnote{https://huggingface.co/datasets/huggan/ukiyoe2photo}, where each dataset inherently contains natural and real style images.
We randomly select a total of $300$ data samples from natural images to perform image style transfer, and similarly choose another $300$ from real style images to compute the FID scores~\cite{fid}.

\paragraph{Baselines.}
To evaluate the effectiveness of style transfer between source images and target styles, we compare our method with several baseline approaches, as follows:
\begin{itemize}
    \item \textit{StyTr}~\cite{Deng2021StyTr2UI} uses dual Transformer encoders and a shared decoder to separately model content and style features, and further introduces a scale-invariant positional encoding to improve robustness in style transfer.
    \item \textit{InstructPix2Pix}~\cite{Brooks2022InstructPix2PixLT} leverages instruction-image pairs to train a text-conditioned diffusion model for fine-grained image editing, and thus enables controllable generation without paired supervision by combining GPT-generated instructions and Stable Diffusion outputs.
    \item \textit{Z*}~\cite{Deng2023ZZS} proposes a zero-shot style transfer strategy by rearranging cross-attention layers in pre-trained diffusion models, where it enables direct style injection by manipulating latent features during denoising in a training-free manner.

    \item \textit{Puff-Net}~\cite{Zheng2024PuffNetES} designs a lightweight Transformer-based model for efficient style transfer by decoupling content and style features, and achieves competitive performance with reduced computation through encoder-only architecture and explicit feature fusion.    
\end{itemize}

\paragraph{Evaluation Metrics.}
Following the conventions of previous style transfer studies~\cite{Hartley2024DomainTS,Song2024ArbitraryMS,Hamazaspyan2023DiffusionEnhancedPA}, we aim to measure the experimental results through automatic metrics, from three main aspects, namely \textit{style consistency}, \textit{content fidelity}, and \textit{aesthetic quality}.
\textit{For style consistency,}
We compute the Fréchet Inception Distance (FID) \cite{fid} between the stylized images and the selected ones from all three benchmarks, where this metric reflects the consistency between their feature distributions. Lower FID values indicate better style transfer performance.
\textit{(ii) Content Fidelity.}
We compute SSIM and LPIPS between the stylized and the source input images to evaluate content fidelity. SSIM measures low-level structural similarity, while LPIPS captures perceptual consistency from deep features. Higher SSIM and lower LPIPS values indicate better content preservation.
\textit{(iii) Aesthetic Quality.}
We use the Aesthetic Score, based on CLIP-extracted features, to assess the aesthetic quality of stylized images. Higher scores indicate better visual appeal.

\begin{figure*}[!t]
\centering
\includegraphics[width=\textwidth]{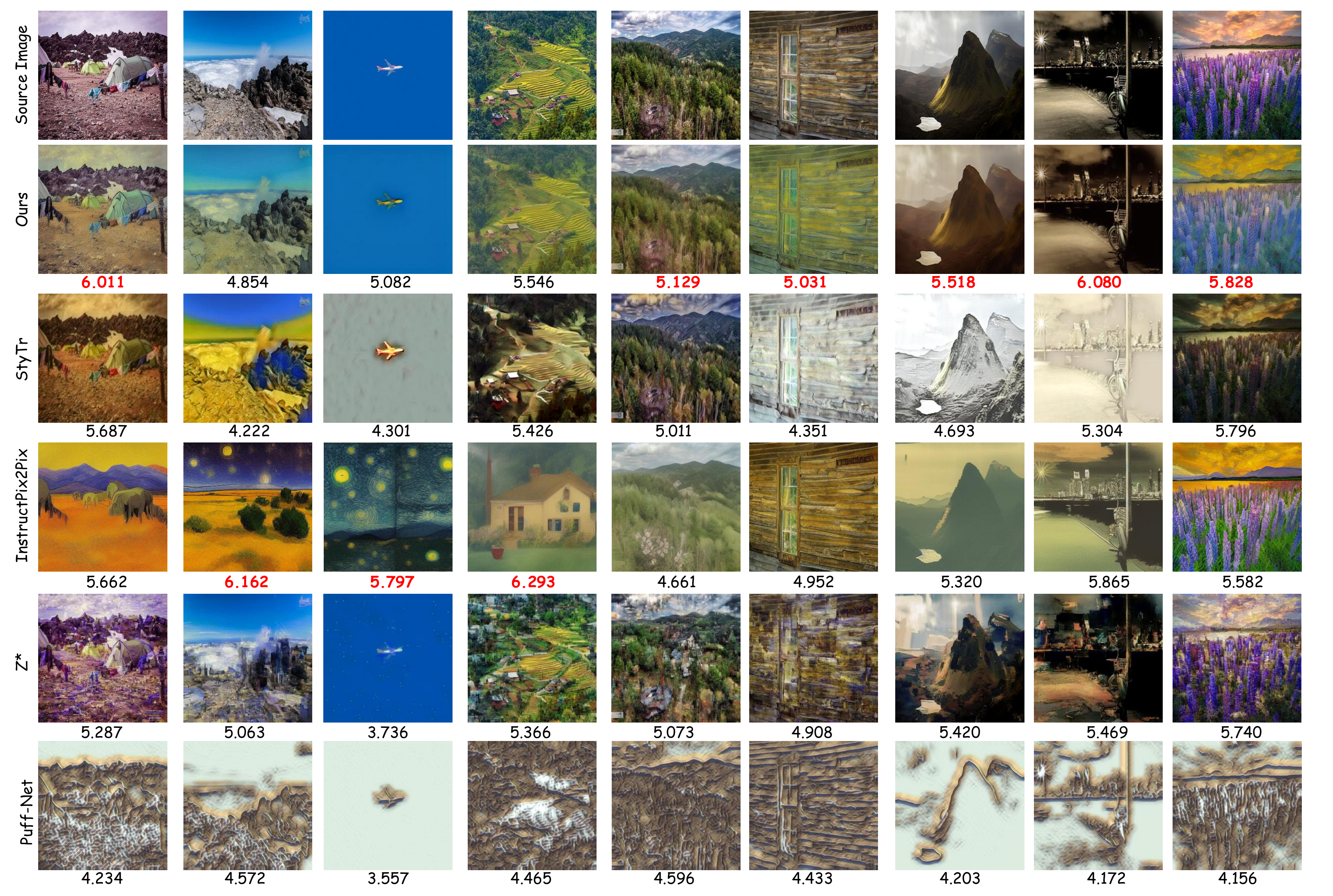}
\caption{
Qualitative comparison of our method against state-of-the-art methods \cite{Deng2021StyTr2UI, Brooks2022InstructPix2PixLT, Deng2023ZZS, Zheng2024PuffNetES}.
Each row displays the stylized results from a specific method, with the corresponding CLIP aesthetic score shown under each image.
Our approach frequently achieves the highest aesthetic scores, marked in red, which reflects a strong balance between the preservation of content and effective stylization.
We highlight the best-performing CLIP aesthetic scores in red.
}
\label{fig:experiment1}
\end{figure*}

\subsection{Quantitative Comparison}
%
Table~\ref{tab:combined_results_all} presents the quantitative comparison compared to existing baseline methods, i.e., StyTr~\cite{Deng2021StyTr2UI}, InstructPix2Pix~\cite{Brooks2022InstructPix2PixLT}, Z*~\cite{Deng2023ZZS}, and Puff-Net~\cite{Zheng2024PuffNetES}, with respect to FID, SSIM, LPIPS, and CLIP aethestic scores.
Specifically in this comparison, StyTr refers to a representative study of image encoding-based architecture, which utilizes a dual Transformer encoders to input the style condition; InstructPix2Pix serves as a more general solution for image-to-image translation and utilize natural language instruction as the hints for model to perform stylization; Z* is a training-free baseline that entirely relies on the generative capabilities of the foundation diffusion model; Puff-Net is similar to StyTr that both employs dual Transformer encoders, where we detail the analyses in the following texts.

Our method demonstrates superior performance over StyTr, achieving a significantly higher SSIM of $0.813$ versus $0.612$ and a lower LPIPS of $0.325$ versus $0.413$ on the Van Gogh dataset, confirming that our method is capable of producing structually well-preserved results than other methods.
This indicates that our model produces images that are not only stylistically faithful but also more structurally aligned with the input content, showing the superior content-style disentangling ability behind.
When compared with InstructPix2Pix, our method shows significant improvements in content preservation, supported by the increase in the SSIM from $0.398$ to $0.813$ and the decrease in the LPIPS from $0.435$ to $0.325$.
This addresses a key limitation where InstructPix2Pix may introduce excessive texture or style inconsistencies, whereas our method maintains a coherent visual structure even in difficult regions like faces or natural objects.
Our approach achieves a more effective trade-off between the fidelity of content and the transfer of style compared to Z*.
While Z* may exhibit more aggressive stylization, it often does so at the expense of semantic consistency, a crucial attribute for the synthesis of high-quality style images.
This highlights that our method prioritizes perceptual fidelity over superficial stylization, as the significant gains in metrics of content preservation often outweigh the small gaps in the scores of aesthetics.
We observe that while some baselines achieve marginally higher aesthetic scores, their generated results can suffer from noticeable structural distortions or inconsistent styles, where we would further detail this phenomena in the next paragraph.

\subsection{Qualitative Comparison}
We conduct qualitative comparison between our method and current state-of-the-art ones in Figure \ref{fig:experiment1}.
Particularly in this comparison, we observe that automatic metrics (particularly in CLIP-ae) might not be well-aligned with visual quality, and thus, we report the CLIP-ae score of each generated sample in Figure \ref{fig:experiment1}.
There are a few observations from various perspectives.
Although StyTr can capture the general color palette, it often fails to preserve the structural integrity of the content.
For instance, in the eighth column, the detailed cityscape is reduced to a simplistic line drawing, which loses essential visual information.
Similarly, the mountain peak in the second column is rendered with unnatural color blocks, failing to convey the texture of the original rock.
InstructPix2Pix frequently achieves high aesthetic scores, yet this often comes at the cost of the fidelity of the content.
As seen in the fourth column, the house is significantly distorted to fit the target style, which sacrifices the structural identity of the original subject.
This tendency to over-stylize leads to the introduction of artistic textures that can overwhelm and lose the semantic structure of the image.
The training-free method Z* demonstrates a tendency towards aggressive stylization that can compromise the clarity of the content.
In the third column, the airplane becomes almost indistinguishable from the background, indicating a significant loss of the semantic information of the object.
Furthermore, the results in the sixth and seventh columns show that the original structures are often replaced by abstract or distorted patterns, which confirms the trade-off for lower semantic consistency.
Puff-Net struggles significantly with preserving content, as its outputs are dominated by repetitive textural artifacts.
Across all examples in its corresponding row, the underlying content is obscured, rendering the images stylistically monotonous and semantically empty.
Conversely, our method maintains a strong balance between the preservation of content and effective stylization.

\begin{figure}[!t]
\centering
\includegraphics[width=\linewidth]{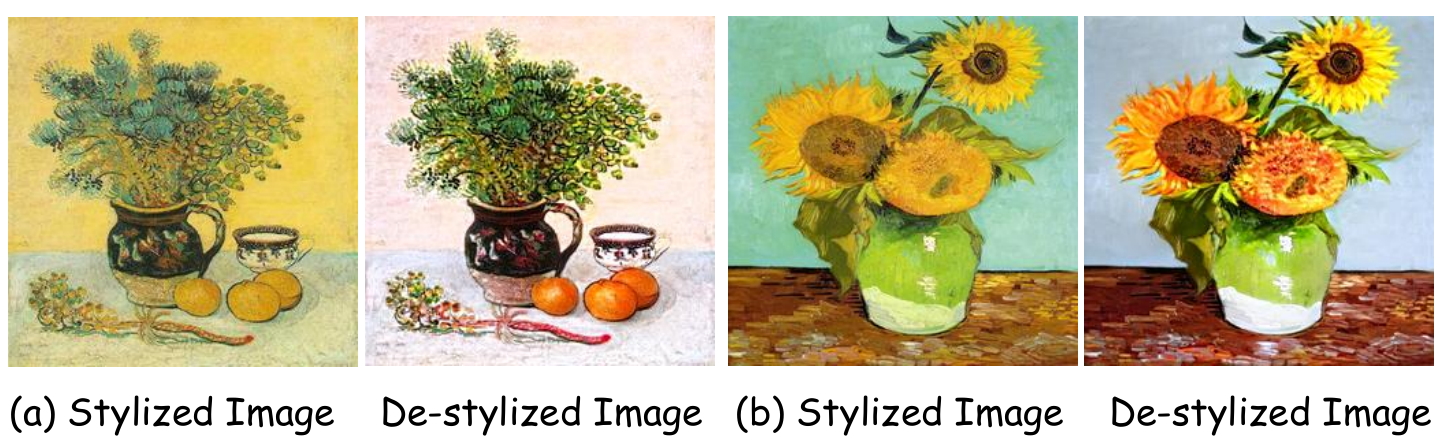}
\caption{
Qualitative results under image de-stylization, where our method is capable of robustly removing stylized artifacts into photorealistic natural images.
}
\label{fig:experiment3}
\end{figure}

Additionally, beyond standard style transfer, our method uniquely supports the task of image de-stylization.
As shown in Figure~\ref{fig:experiment3}, our approach effectively inverts the stylization process by removing the characteristic artistic features while restoring a natural appearance.
For instance, in example (a), the model strips away the heavy brushstrokes and the yellow-tinted palette typical of the style of Van Gogh.
It concurrently restores photorealistic details, such as the distinct textures of the fruit and the foliage, while precisely preserving the overall composition.
Similarly, in the case of the sunflowers (b), the iconic impasto is replaced with a rendering that features naturalistic lighting and smoother surfaces on the petals and the vase.
Both aforementioned results confirm the robust ability of our method in de-stylizing images.





\subsection{Ablation Studies}

In this section, to validate the effectiveness of the key components in our proposed framework, we conduct ablation studies from two aspects: different training strategies and the choice of generators used for data distillation. Due to limited computational resources, all experiments are performed on the VanGogh2Photo dataset.

\begin{table}[!t]
\centering
\small 
\setlength{\tabcolsep}{0.3em}
\begin{tabular}{
    l @{\hskip 16pt}
    c @{\hskip 14pt}
    c @{\hskip 14pt}
    c @{\hskip 4pt}
    c
}
\toprule
Method & FID$\downarrow$ & SSIM$\uparrow$ & LPIPS$\downarrow$ & CLIP-ae$\uparrow$ \\
\midrule
Two-Stage Training & 195.9 & \textbf{0.824} & \underline{0.314} & 5.701 \\
Joint Training (Ours) & \textbf{188.2} & 0.813 & 0.325 & 5.705 \\
\bottomrule
\end{tabular}
\caption{Ablation study on the training strategy. We compare our end-to-end \textit{Joint Training} approach with a \textit{Two-Stage Training} baseline. Best results from the overall ablation are in \textbf{bold}, second-best are \underline{underlined}.}
\label{tab:ablation_training_strategy}
\end{table}

\begin{table}[!t]
\centering
\setlength{\tabcolsep}{0.12em}
\small 
\begin{tabular}{
    l @{\hskip 16pt}
    c @{\hskip 14pt}
    c @{\hskip 14pt}
    c @{\hskip 4pt}
    c
}
\toprule
Method & FID$\downarrow$ & SSIM$\uparrow$ & LPIPS$\downarrow$ & CLIP-ae$\uparrow$ \\
\midrule
Flux-dev Distillation & \underline{195.6} & 0.801 & \textbf{0.129} & \underline{5.821} \\
InstructPix2Pix (Ours) & \textbf{188.2} & 0.813 & 0.325 & 5.705 \\
\bottomrule
\end{tabular}
\caption{Ablation study on the data distillation generator. We compare using \textit{Flux-dev} against our default choice, \textit{InstructPix2Pix}. Best results from the overall ablation are in \textbf{bold}, second-best are \underline{underlined}.}
\label{tab:ablation_distillation_generator}
\end{table}

\begin{table}[!t]
\centering
\small
\begin{tabular}{
    l @{\hskip 16pt}
    c @{\hskip 14pt}
    c @{\hskip 14pt}
    c @{\hskip 4pt}
    c
}
\toprule
Method & FID$\downarrow$ & SSIM$\uparrow$ & LPIPS$\downarrow$ & CLIP-ae$\uparrow$ \\
\midrule
Without LoRA & 196.5 & \underline{0.821} & 0.325 & \textbf{5.818} \\
With LoRA (Ours) & \textbf{188.2} & 0.813 & 0.325 & 5.705 \\
\bottomrule
\end{tabular}
\caption{Ablation study on the effect of LoRA modules. We compare our full model against a variant with all LoRA modules removed. Best results from the overall ablation are in \textbf{bold}, second-best are \underline{underlined}.}
\label{tab:ablation_lora_effect}
\end{table}

\paragraph{Two-Stage Training \textit{vs.} Joint Training.}
To evaluate the impact of the training methodology, we compare our end-to-end joint training scheme with a two-stage sequential approach.
In the two-stage method, the LoRA modules for the UNet and VAE are trained first, followed by the training of the LoRA modules for the text encoder.
As presented in Table~\ref{tab:ablation_training_strategy}, our joint training approach achieves a substantially lower FID of $188.2$ compared to $195.9$ from the two-stage method, which indicates a significant improvement in the fidelity of style.
Although the two-stage method yields marginally better scores for SSIM and LPIPS, the visual results shown in Figure~\ref{fig:experiment3} reveal its limitations.
The two-stage approach tends to suffer from style drift or over-smoothing, failing to preserve key structural details like the contours of the camel.
These findings suggest that end-to-end optimization brings a more effective synergy between the textual and visual branches of the model.
This joint optimization resolves semantic and stylistic inconsistencies by enabling mutual adaptation between the language and vision components, leading to a more coherent representation space that benefits both the fidelity of style and content.

%
%
%

\paragraph{Different Generators in Data Distillation.}
We investigate the impact of the data distillation source by comparing two generators used for style data synthesis: \textit{InstructPix2Pix} and \textit{Flux-dev}.
An analysis of the quantitative results in Table~\ref{tab:ablation_distillation_generator} reveals a notable trade-off.
While using data from Flux-dev yields a lower LPIPS of $0.129$ and a higher aesthetic score of $5.821$, our chosen generator InstructPix2Pix achieves a superior FID of $188.2$ versus $195.6$.
However, the visual results presented in Figure~\ref{fig:experiment3} clearly demonstrate the limitations of relying solely on these metrics.
The output from the model trained on data from Flux-dev suffers from significant style drift, resulting in overly smoothed textures that fail to capture the specified artistic style.
We attribute this outcome to the effective instruction-following capability of InstructPix2Pix, which allows it to distill high-quality and style-specific data.
This underscores the critical importance of the quality and diversity of distilled data in unsupervised training, as this foundation directly influences the expressiveness and fidelity of the final stylized outputs.

\begin{figure}[!t]
\centering
\includegraphics[width=\linewidth]{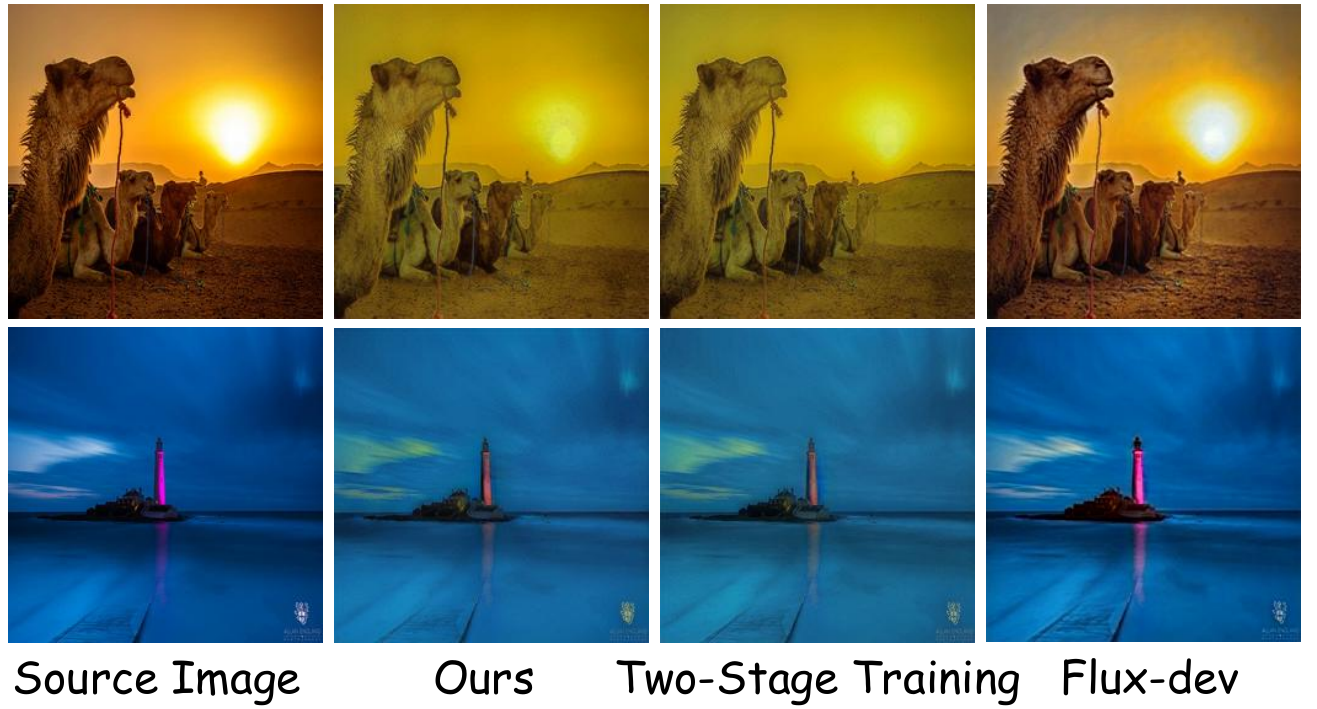}  
\caption{
Qualitative results for all ablation studies on the two-stage training, different generators in data distillation, and the effect of LoRA modules.
Here, ``Two-Stage Training'' refers to the baseline using a two-stage training strategy rather than a joint training one; ``Flux-dev'' denotes the method variant that uses text-to-image FLUX model in simulating stylized images; ``Ours'' represents the full model of our proposed method.
%
}
\label{fig:experiment2}
\end{figure}

\paragraph{Effect of LoRA Modules.}
LoRA weights strongly correlate to the impact of our distribution-specific learning strategy in CSD, where it is vital to investigate its effectiveness.
Therefore, to validate the effectiveness of it, we conduct an experiment by ablating the LoRA modules of the UNet, VAE, and text encoder.
The quantitative results in Table~\ref{tab:ablation_lora_effect} highlight the critical role of LoRA.
The removal of LoRA modules causes a significant degradation in the fidelity of style, as evidenced by the increase in the FID from $188.2$ to $196.5$.
This confirms that the LoRA modules are crucial for improving not only the fidelity of style but also the overall visual expressiveness and diversity of the results, which offers distribution-specific weights to learn paticular patterns from large-scale, unsupervised data.

%
%


\section{Conclusion}

In this paper, we introduce StyDeco, a novel unsupervised framework that resolves a fundamental conflict in text-guided image style transfer, i.e., the semantic-spatial misalignment between monolithic text prompts and the spatially-aware nature of visual style.
Our approach first employs PGD to automatically synthesize a pseudo-paired dataset from a frozen generative model.
Subsequently, we introduce a Contrastive Semantic Decoupling objective that learns distinct text representations for the content and style domains.
Extensive experiments on three typical style patterns of Van Gogh, Cezanne, and Ukiyoe demonstrate that our method demonstrate promising performance in image style transfer, with further analyses illustrating the internal impacts of different components.
Our framework offers a new and effective paradigm for unsupervised image generation, providing a robust solution for text-controllable image-to-image translation, where we wish this work might serve as a reference frameworks for follow-ups in the future.

\bibliography{aaai2026}

\end{document}